\pdfoutput=1

\documentclass[11pt]{article}

\usepackage[final]{acl}

\usepackage{times}
\usepackage{latexsym}

\usepackage[T1]{fontenc}

\usepackage[utf8]{inputenc}

\usepackage{microtype}

\usepackage{inconsolata}

\usepackage{graphicx}

\usepackage{microtype}
\usepackage{hyperref}
\usepackage{url}
\usepackage{booktabs}
\usepackage{graphicx}
\usepackage{wrapfig}
\usepackage{mathtools}
\usepackage{amsmath}
\usepackage{breqn}
\usepackage{latexsym}
\usepackage{lipsum}
\usepackage{enumerate}
\usepackage{enumitem}

\definecolor{darkblue}{rgb}{0, 0, 0.5}
\hypersetup{colorlinks=true, citecolor=darkblue, linkcolor=darkblue, urlcolor=darkblue}
\definecolor{darkpastelred}{rgb}{0.76, 0.23, 0.13}
\definecolor{coolgrey}{rgb}{0.55, 0.57, 0.67}
\definecolor{lowyellow}{RGB}{241, 196, 15}

\title{ParaICL: Towards Parallel In-Context Learning}

\author{
Xingxuan Li$^{1,2}$\thanks{Work done while at DAMO Academy, Alibaba Group. Xingxuan Li was under the Joint Ph.D. Program between Alibaba and Nanyang Technological University.}~
Xuan-Phi Nguyen$^{3}$\footnotemark[1]~
Shafiq Joty$^{2,3}$~
Lidong Bing$^{1}$\footnotemark[1]\thanks{Corresponding author.}\\
$^1$Shanda AI Research Institute $^2$Nanyang Technological University $^3$Salesforce Research\\
\{xingxuan.li, lidong.bing\}@shanda.com
\{nguyenxu002, srjoty\}@ntu.edu.sg ~
}

\begin{document}
\maketitle

\begin{abstract}
Large language models (LLMs) have become the norm in natural language processing (NLP), excelling in few-shot in-context learning (ICL) with their remarkable abilities.
Nonetheless, the success of ICL largely hinges on the choice of few-shot demonstration examples, making the selection process increasingly crucial.
Existing methods have delved into optimizing the quantity and semantic similarity of these examples to improve ICL performances.
However, our preliminary experiments indicate that the effectiveness of ICL is limited by the length of the input context.
Moreover, varying combinations of few-shot demonstration examples can significantly boost accuracy across different test samples.
To address this, we propose a novel method named parallel in-context learning (ParaICL) that effectively utilizes all demonstration examples without exceeding the manageable input context length. 
ParaICL employs parallel batching to distribute demonstration examples into different batches according to the semantic similarities of the questions in the demonstrations to the test question. 
It then computes normalized batch semantic scores for each batch. 
A weighted average semantic objective, constrained by adaptive plausibility, is applied to select the most appropriate tokens. 
Through extensive experiments, we validate the effectiveness of ParaICL and conduct ablation studies to underscore its design rationale.
We further demonstrate that ParaICL can seamlessly integrate with existing methods.
Our code is available at \href{https://github.com/xingxuanli/paraicl.git}{https://github.com/xingxuanli/paraicl.git}.
\end{abstract}

\section{Introduction}
\label{sec:intro}
In recent years, scaling up the parameters of generative language models has significantly enhanced their language generation capabilities \citep{Radford2019LanguageMA, brown2020language, openai2023gpt4}.
Large language models (LLM) have demonstrated their adeptness across a wide range of tasks via few-shot in-context learning (ICL) \citep{cheng2023gpt4,zhao2023verifyandedit, li2024cok}. 
In few-shot ICL, models are expected to generate outputs directly based on a sequence of given examples without any parameter modifications, making it the most efficient method for adapting to new tasks.

\begin{figure}[t!]
    \centering
    \includegraphics[width=1\linewidth]{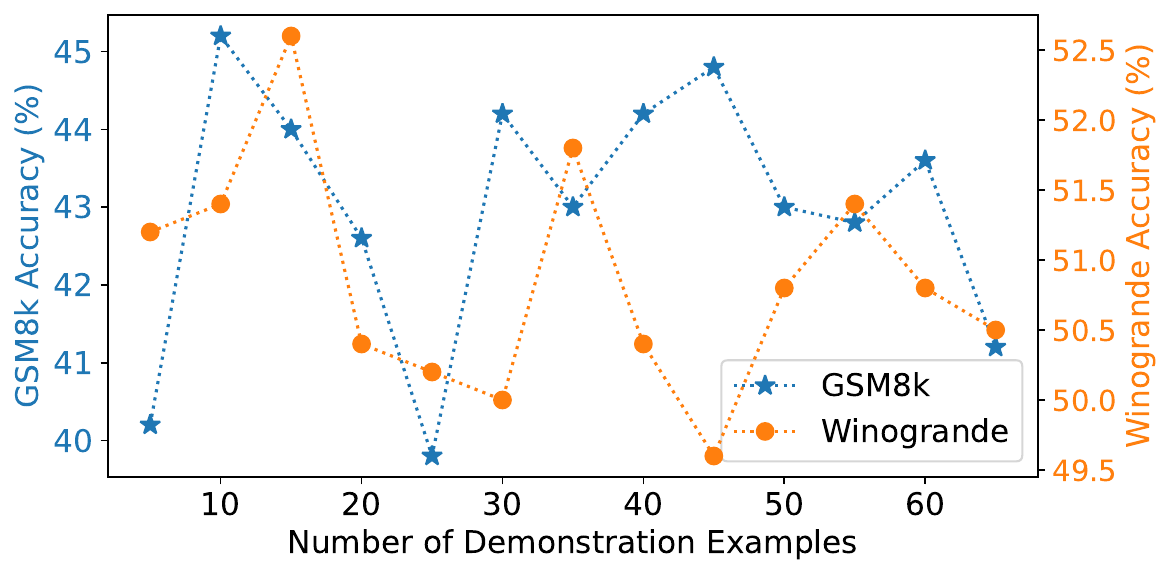}
    \caption{Results of Mistral-7B-Instruct-v0.2 on 100 test samples from GSM8K and WinoGrande using different numbers of few-shot demonstration examples. Increasing the number of demonstration examples does not necessarily improve the performance consistently.}
    \label{fig:pre_1}
\end{figure}

However, the effectiveness of ICL is notably influenced by the few-shot demonstration examples used \citep{chen-etal-2023-many}.
Various methods have been developed to select the most effective few-shot demonstration examples for ICL.
\citet{hao2022structured} demonstrated that scaling up the number of demonstration examples can improve the ICL performance.
\citet{rubin2022learning} and \citet{liu2021makes} proposed to utilize the most relevant examples for each test sample during inference.
Nevertheless, existing few-shot ICL methods mostly focus on either increasing the number of demonstration examples or selecting a few that are similar to the test samples to improve performance.

We have designed two preliminary experiments to identify key factors for effectively using demonstration examples.
Firstly, we increase the number of demonstration examples.
The results, as shown in Figure \ref{fig:pre_1}, reveal that the performance of Mistral-7B-Instruct-v0.2 on GSM8K and WinoGrande does not consistently improve with more examples.
This is partly because longer input lengths, resulting from more examples, can lead to suboptimal results in LLMs \citep{liu2023lost, li2023unlocking}.
Therefore, controlling the input context length is essential in ICL, making the number of few-shot demonstration examples a critical factor.

\begin{figure}[t!]
    \centering
    \includegraphics[width=\linewidth]{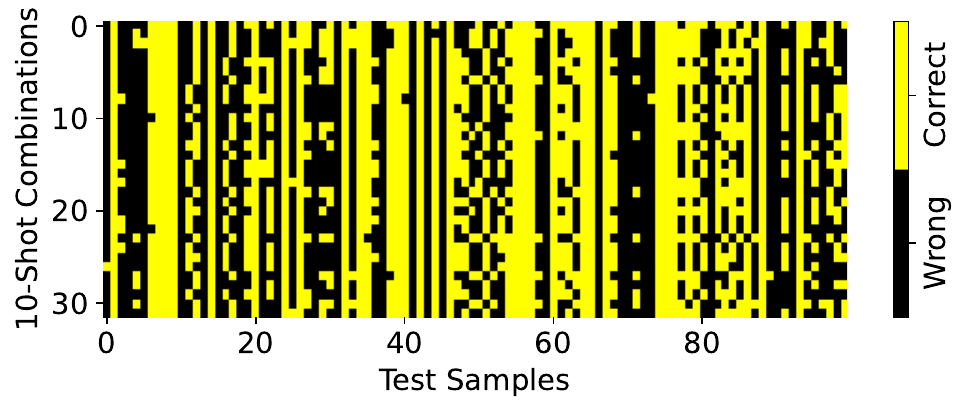}
    \caption{Results of Llama-2-7B-Chat on 100 WinoGrande test samples using different combinations of 10-shot demonstration examples. Different combinations improve the model's accuracy on various test samples.}
    \label{fig:pre_2}
\end{figure}

Secondly, we focus on the selection of demonstration examples.
We experiment with Llama-2-7B-Chat on 100 test samples from WinoGrande with 32 different 10-shot example combinations.
As shown in Figure \ref{fig:pre_2}, we notice that varying combinations lead to different accuracy improvements.
These combinations enabled the model to accurately answer 80\% of the test samples, a significant increase compared to the 50.9\% average accuracy of each individual 10-shot combination.
This highlights the fact that varying demonstration examples can enhance the model's accuracy on different test samples.
Therefore, we should leverage all available demonstration examples when possible.

To combine the best of both worlds, we introduce parallel in-context learning (ParaICL).
ParaICL effectively utilizes the maximum number of demonstration examples without extending the input context length, thus avoiding the potential reduction in model performance due to larger context sizes.
Given a set of question--answer pairs as demonstration examples, ParaICL first assigns them into various batches based on the semantic similarity between the demonstrations' questions and the sample test question.
Consequently, each batch maintains a controlled context length while utilizing all demonstration examples.
These batches are then processed by a causal model in parallel to obtain the next token distribution for each batch.
Afterward, a weighted average of these distributions is calculated, considering the semantic relation of each batch to the test question.
The final step involves selecting the token with the highest weighted average probability for continued generation.

We conduct extensive experiments across a range of reasoning, natural language inference, and coding tasks to validate the effectiveness of ParaICL.
We further demonstrate that ParaICL is compatible with both open- and closed-source causal language models.
Our study includes abundant ablation studies and analyses to justify the design of ParaICL and demonstrate how it can be integrated with other ICL methods.
In summary, our main contributions are the following:
(1) We introduce ParaICL, a simple but effective method that leverages all available demonstration examples while maintaining the input context length manageable.
(2) We conduct thorough experiments to prove the effectiveness of our method, along with ablation studies to justify its design.
(3) We illustrate how our method can enhance and work in conjunction with other methods.

\section{Related Work}
ICL has surged as a transformative approach in the NLP domain, with its significance evident across various applications, including knowledge grounding \citep{zhao2023verifyandedit, li2024cok}, code generation \citep{li2023unlocking}, and other industrial applications \citep{cheng2023gpt4, chen2024chatgpts}.
Despite its promise, challenges such as sensitivity to the prompts and context window length limitations, highlight areas for further research and development in making ICL more robust and versatile.

As aforementioned, various studies have underscored that ICL exhibits significant sensitivity to the quality of prompts, particularly concerning the demonstration examples provided \citep{li-qiu-2023-finding, gupta-etal-2023-coverage, bayesian-icl}.
\citet{chen-etal-2023-many} discovered that increasing the number of examples only leads to slight improvements.
\citet{liu2021makes} proposed to select examples that have the highest semantic similarity to the test question.
Contrarily, \citet{levy2023diverse} found that leveraging a diverse set of demonstration examples could improve in-context compositional generalization.
\citet{qin2023incontext} proposed iterative demonstration selection, which considers both the diversity and similarity dimensions of ICL demonstration selection for LLMs.
Additionally, \citet{chia2023contrastive} explored the potential of contrastive examples in improving the reasoning capabilities of causal language models.
Nevertheless, these methods primarily focus on choosing a subset of demonstration examples from a pool of candidates based on certain perspectives.
Furthermore, these methods necessitate a substantial number of candidates to efficiently select the best examples, which constrains their applicability across all situations.

Another line of research is dedicated to utilizing extended context in ICL.
\citet{ratner2023parallel} proposed parallel context window (PCW), a method that alleviates the context window limitations for any open-source LLMs without necessitating additional training.
However, PCW overlooked the semantic connections between the demonstration examples and the test samples.
As evidenced by \citet{chen-etal-2023-many}, merely increasing the quantity of demonstration examples does not correlate with a substantial improvement in performance.
In fact, this strategy might detract from overall performance due to the inclusion of incorrect or misleading content within some of the examples.

The implementation of our ParaICL method is also inspired by the ensemble of models \citep{Ganaie_2022}, which use multiple slightly different models to contribute probability ``votes'' before producing the final answer by (weighted) averaging the votes' probabilities. Vastly different from the ensemble of models, ParaICL uses the same language model prompted with different sets of in-context exemplars to produce varying vote probabilities before aggregating the output distributions to produce the final answer. As such, ParaICL does not require many models and is efficient to compute.

\section{Methodology}
\begin{figure*}[t]
    \centering
    \includegraphics[width=0.95\textwidth]{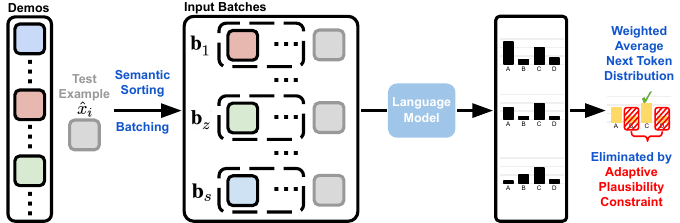}
    \caption{Our proposed parallel in-context learning (ParaICL) method. Colored squares with \textbf{black} borders denote demonstration samples. Squares filled in \textcolor{coolgrey}{\textbf{grey}} with matching borders denote test sample $\hat{x}_i$.}
    \label{fig:method}
\end{figure*}

We first formulate the problem setting for few-shot ICL in Section \ref{sec:icl}.
Following this, in Section \ref{sec:batching}, we introduce ParaICL, a novel method that efficiently processes demonstration examples in batches and effectively utilizes a novel parallel semantic decoding strategy for generation.
Subsequently in Section \ref{sec:psd}, we introduce the weighted average semantic objective alongside the adaptive plausibility constraint.
Finally we define the parallel semantic decoding method, which employs the weighted average semantic objective subject to the plausibility constraint.
A demonstration of ParaICL can be found in Figure \ref{fig:method}.

\subsection{Few-Shot In-Context Learning}
\label{sec:icl}
Few-shot ICL focuses on understanding and executing tasks with a set of demonstration examples \citep{brown2020language}.
It leverages a number of selected examples, known as ``shots'', to quickly adapt to new tasks.
Specifically, within the framework of $k$-shot ICL, the model is provided with $k$ demonstration examples, represented as $\mathcal{D}=\{(x_1,y_1),...,(x_k,y_k)\}$, incorporated into the input prompt for context.

\subsection{Parallel Batching}
\label{sec:batching}
The preliminary experiments in Section \ref{sec:intro} demonstrate the importance of employing varied combinations of demonstration examples without extending the length of the input context.
Consequently, we proceed by organizing the demonstration examples $\mathcal{D}=\{(x_1,y_1),...,(x_k,y_k)\}$ into batches.
Previous studies have shown that the selection of semantically significant demonstration examples enhances ICL performances \citep{liu2021makes, luo2023dricl}.
Therefore, for each test question $\hat{x}_i \in \mathcal{\hat{D}}=\{(\hat{x}_1,\hat{y}_1),...,(\hat{x}_n,\hat{y}_n)\}$, we first sequence the demonstration examples in $\mathcal{D}$ by their question semantic similarities to the test question $\hat{x}_i$.
This results in an ordered sequence $\mathcal{D}_{sorted}^{i}=\{(x_{(1)}^{i},y_{(1)}^{i}),...,(x_{(k)}^{i},y_{(k)}^{i})\}$, where the similarity function $f_{sim}$ determines the order based on the cosine similarity of the input embeddings:
\begin{equation}
    f_{sim}(x_{(1)}^{i},\hat{x}_i) \geq ... \geq f_{sim}(x_{(k)}^{i},\hat{x}_i).
\end{equation}
$f_{sim}$ is formulated as below,
\begin{equation}
    f_{sim}(t_1,t_2) = \frac{f_{emb}(t_1)\cdot f_{emb}(t_2)}{ \lVert f_{emb}(t_1) \rVert \lVert f_{emb}(t_2)\rVert},
\end{equation}
where $t_1$ and $t_2$ are the input texts, and $f_{emb}(\cdot)$ is a model to compute the sentence semantic embedding of the input text.

With $m$ defined as the divisor of $k$ that indicates the number of examples per batch, we form $s$ batches, where $s = \frac{k}{m}$.
These sorted demonstration examples $\mathcal{D}^{i}_{sorted}$ are then divided into $s$ parallel batches, denoted as $\mathcal{B}^{i}=\{\mathrm{\textbf{b}}_1^{i},...,\mathrm{\textbf{b}}_z^{i},...,\mathrm{\textbf{b}}_{s}^{i}\}$, with each $\mathrm{\textbf{b}}_z^{i}$ including demonstration examples $\{(x_{((z-1)m+1)}^{i}, y_{((z-1)m+1)}^{i}),...,(x_{(zm)}^{i}, y_{(zm)}^{i})\}$, where $1 \leq z \leq s$.

For each batch, we calculate normalized batch similarity scores $\mathcal{O}^{i} = \{o_1^{i},...,o_z^{i},...,o_{s}^{i}\}$, where each $o_z^{i}$ represents the semantic similarity of the batch's demonstration questions to the test question, determined by:
\begin{equation}
    o_z^{i} = \frac{\sum_{r=1}^m{f_{sim}(x_{((z-1)m+r)}^{i}, \hat{x}_i)}}{\sum_{z=1}^{s}{\sum_{r=1}^m{f_{sim}(x_{((z-1)m+r)}^{i}, \hat{x}_i)}}}.
\end{equation}
These scores measure the semantic similarity between the demonstration examples in each batch and the test question.

Finally, we compile the input prompts for each batch by incorporating the test question, creating $\mathcal{U}^{i}=\{\mathrm{\textbf{u}}_1^{i},...,\mathrm{\textbf{u}}_z^{i},...,\mathrm{\textbf{u}}_{s}^{i}\}$, where $\mathrm{\textbf{u}}_z^{i}=\{\mathrm{\textbf{b}}_z^{i},\hat{x_i}\}$.

\subsection{Parallel Semantic Decoding}
\label{sec:psd}
We first define a generative language model $f_{lm}(\cdot)$. 
For given inputs $\mathrm{\textbf{u}}_{prev}^{i}$, the model generates a continuation $\mathrm{\textbf{u}}_{cont}^{i}$ according to the formula:
\begin{equation}
    f_{lm}(\mathrm{\textbf{u}}_{cont}^{i}|\mathrm{\textbf{u}}_{prev}^{i}) = \prod_{j=1}^{q^{i}}{f_{lm}(u_j^{i}|u_{<j}^{i},\mathrm{\textbf{u}}_{prom}^{i})},
\end{equation}
where $u_j^{i}$ is a generated token in $\mathrm{\textbf{u}}_{cont}^{i}$, $q^{i}$ represents the total number of generated tokens, and $\mathrm{\textbf{u}}_{prom}^{i}$ is the provided input prompt.

\paragraph{Weighted average semantic objective}
As demonstration examples have varying semantic contributions to the test sample, we propose the weighted average semantic (WAS) objective.
The WAS objective for test sample $\hat{x}_i$ is defined as:
\begin{dmath}
    \mathcal{L}_{\mathrm{WAS}}(\mathrm{\textbf{u}}_{cont}^{i}, \mathcal{O}^{i}, \mathcal{U}^{i}) = \sum_{z=1}^{s}{o_z^{i} \cdot f_{lm}{(\mathrm{\textbf{u}}_{cont}^{i}|\mathrm{\textbf{u}}_z^{i})}}.
\end{dmath}
The WAS objective rewards batches with demonstration examples that exhibit greater semantic similarity to the test question, while simultaneously leveraging all examples to enrich the generation process comprehensively.
This advantage makes ParaICL superior to PCW, which arranges examples randomly.
However, certain batches may contain noise that could adversely affect the performance.
To address this problem, we adopt the adaptive plausibility constraint from the contrastive decoding method \citep{li2023contrastive}.

\paragraph{Adaptive plausibility constraint}
The adaptive plausibility constraint $\mathcal{V}_{head}$ leverages the confidence level in the foremost batch to mitigate the impact of potentially less relevant demonstration batches.
Explicitly, the adaptive plausibility constraint for test sample $\hat{x}_{i}$ is defined as:
\begin{equation}
\begin{split}
        & \mathcal{V}_{head}(u_{<j}^{i}) \\
    & = \{ u_j^{i} \in \mathcal{V} : f_{lm}(u_j^{i}|u_{<j}^{i}, \mathrm{\textbf{u}}_1^{i}) \\
    & \quad\quad \ge \alpha \max_{w^{i}} f_{lm} (w^{i} | u_{<j}^{i}, \mathrm{\textbf{u}}_1^{i}) \},
\end{split}
\end{equation}
where $\mathcal{V}$ represents the model's vocabulary, $\mathrm{\textbf{u}}_1^{i}$ is the first batch (\textit{i.e.,} the most semantically aligned with the test question $\hat{x}_i$), and $\alpha$ is a hyperparameter between $[0,1]$ \footnote{Following the contrastive decoding method \citep{li2023contrastive}, we set $\alpha$ as 0.1 for all experiments.}.
A higher $\alpha$ value signifies a preference for tokens with higher generation probabilities, whereas smaller $\alpha$ allows tokens of lower probabilities to be generated.

\paragraph{Final method}
The final parallel semantic decoding method combines the WAS objective with the adaptive plausibility constraint to optimize the generation process:
\begin{equation}
    \begin{split}
        & \max_{\mathrm{\textbf{u}}_{cont}^{i}}{\mathcal{L}_{\mathrm{WAS}}(\mathrm{\textbf{u}}_{cont}^{i}, \mathcal{O}^{i}, \mathcal{U}^{i})} \\
        & \text{subject to } u_j^{i} \in \mathcal{V}_{head}(u_{<j}^{i}), \forall u_j^{i} \in \mathrm{\textbf{u}}_{cont}^{i}.
    \end{split}
\end{equation}

Given the complexity at the sequence level, we simplify the optimization to the token level as follows:
\begin{equation}
    \begin{split}
        & \mathcal{L}_{\mathrm{WAS}}(\mathrm{\textbf{u}}_{cont}^{i}, \mathcal{O}^{i}, \mathcal{U}^{i}) \\
        & = \sum_{z=1}^{s}{o_z^{i} \cdot f_{lm}{(\mathrm{\textbf{u}}_{cont}^{i}|\mathrm{\textbf{u}}_z^{i})}} \\
        & = \prod_{j=1}^{q^{i}}{\text{WAS-score}(u_j^{i}, u_{<j}^{i}, \mathcal{O}^{i}, \mathcal{U}^{i})},
    \end{split}
\end{equation}
where $\text{WAS-score}(u_j^{i}, u_{<j}^{i}, \mathcal{O}^{i}, \mathcal{U}^{i})$ is the token level score formulated as:
\begin{equation}
    \begin{split}
        & \text{WAS-score}(u_j^{i}, u_{<j}^{i}, \mathcal{O}^{i}, \mathcal{U}^{i}) \\
        & =
            \begin{cases}
                \sum\limits_{z=1}^{s}{o_z^{i} \cdot f_{lm}(u_j^{i}|u_{<j}^{i},\mathrm{\textbf{u}}_z^{i})} & \text{if } u_j^{i} \in \mathcal{V}_{head}(u_{<j}^{i}),\\
                -\inf & \text{otherwise}. \\
            \end{cases}
    \end{split}
\end{equation}
We first apply plausibility constraints $\mathcal{V}_{head}(u_{<j}^{i})$ to filter tokens, discarding those that do not reach the required probability threshold within the most semantically pertinent demonstration batch.
Subsequently, the surviving tokens are evaluated using the weighted average semantic scores derived from all batches.
Consequently, this process allows for the selection of a token that incorporates information from every batch of examples.

\section{Experiments}
\begin{table*}[t]
\begin{center}
\resizebox{\textwidth}{!}{
\begin{tabular}{lcccccccccc}
\toprule

&\multicolumn{5}{c}{\textbf{Llama-2-7B-Chat}} & \multicolumn{5}{c}{\textbf{Mistral-7B-Instruct-v0.2}} \\
\cmidrule(lr){2-6}
\cmidrule(lr){7-11}
\textbf{Method} & \textbf{GSM8K} & \textbf{WinoGrande} & \textbf{ARC-C} & \textbf{HellaSwag} & \textbf{MBPP} & \textbf{GSM8K} & \textbf{WinoGrande} & \textbf{ARC-C} & \textbf{HellaSwag} & \textbf{MBPP} \\

\hline
\multicolumn{11}{c}{3-shot} \\ 
\hline
Standard & 19.9 \small{$\pm$ 0.9}                      & 52.7 \small{$\pm$ 1.1}                      & 54.1 \small{$\pm$ 1.2}                      & 32.9 \small{$\pm$ 1.7}                      & 19.5 \small{$\pm$ 0.6}                      & 40.7 \small{$\pm$ 1.9}                      & 58.0 \small{$\pm$ 2.1}                      & 70.5 \small{$\pm$ 0.6}                     & 58.6 \small{$\pm$ 1.2}                      & 30.2 \small{$\pm$ 2.1} \\
Sorted+  & 20.1 \small{$\pm$ 0.8}                      & {53.1 \small{$\pm$ 1.3}}          & 53.9 \small{$\pm$ 0.8}                      & 33.6 \small{$\pm$ 1.1}                      & 18.7 \small{$\pm$ 1.9}                      & 39.2 \small{$\pm$ 1.4}                      & 58.8 \small{$\pm$ 0.7}                      & 72.1 \small{$\pm$ 1.8}                     & 58.9 \small{$\pm$ 1.1}                      & 31.5 \small{$\pm$ 0.8} \\
Sorted-  & 18.4 \small{$\pm$ 1.6}                      & 52.3 \small{$\pm$ 0.6}                      & 54.3 \small{$\pm$ 0.9}                      & 33.2 \small{$\pm$ 0.5}                      & 19.6 \small{$\pm$ 2.4}                      & 40.9 \small{$\pm$ 1.2}                      & 57.7 \small{$\pm$ 1.3}                      & 71.2 \small{$\pm$ 0.9}                     & 58.5 \small{$\pm$ 0.5}                      & 30.6 \small{$\pm$ 0.9} \\
{CBS}  & 17.3 \small{$\pm$ 1.2}                      & {51.9 \small{$\pm$ 1.0}}                      & 52.8 \small{$\pm$ 1.1}                      & 31.5 \small{$\pm$ 1.6}                      & 17.2 \small{$\pm$ 0.7}                      & 38.1 \small{$\pm$ 1.5}                      & 56.6 \small{$\pm$ 1.1}                      & 69.9 \small{$\pm$ 1.4}                     & 57.7 \small{$\pm$ 1.3}                      & 30.9 \small{$\pm$ 1.4} \\
{LENS}  & 16.8 \small{$\pm$ 0.7}                      & {52.5 \small{$\pm$ 0.9}}                      & 53.4 \small{$\pm$ 1.3}                      & 32.1 \small{$\pm$ 0.7}                      & 16.8 \small{$\pm$ 1.8}                      & 39.5 \small{$\pm$ 1.7}                      & 55.3 \small{$\pm$ 1.4}                      & 69.4 \small{$\pm$ 1.8}                     & 57.1 \small{$\pm$ 1.6}                      & 29.3 \small{$\pm$ 1.3} \\
PCW  & 17.6 \small{$\pm$ 1.1}                      & \underline{53.4 \small{$\pm$ 0.7}}                      & 52.2 \small{$\pm$ 0.7}                      & 33.5 \small{$\pm$ 0.8}                      & 16.2 \small{$\pm$ 1.3}                      & 40.5 \small{$\pm$ 0.8}                      & 58.9 \small{$\pm$ 1.2}                      & 70.1 \small{$\pm$ 1.3}                     & 58.2 \small{$\pm$ 0.8}                      & 31.9 \small{$\pm$ 1.7} \\
{SP}  & 19.4 \small{$\pm$ 1.5}                      & {51.6 \small{$\pm$ 1.5}}                      & 51.6 \small{$\pm$ 1.9}                      & 31.9 \small{$\pm$ 1.9}                      & 19.1 \small{$\pm$ 1.5}                      & 37.4 \small{$\pm$ 2.5}                      & 56.1 \small{$\pm$ 1.1}                      & 70.6 \small{$\pm$ 1.1}                     & 57.5 \small{$\pm$ 0.9}                      & 30.4 \small{$\pm$ 0.8} \\
ParaICL  & \underline{22.1 \small{$\pm$ 1.3}}          & 52.9 \small{$\pm$ 0.5}                      & \underline{55.3 \small{$\pm$ 0.6}}          & \underline{33.7 \small{$\pm$ 2.2}}          & \textbf{\underline{20.9 \small{$\pm$ 0.6}}} & \underline{41.1 \small{$\pm$ 1.3}}          & \underline{59.2 \small{$\pm$ 0.9}}          & \underline{71.9 \small{$\pm$ 0.6}}         & \textbf{\underline{59.2 \small{$\pm$ 1.5}}} & \underline{32.6 \small{$\pm$ 1.1}} \\
\hline
\multicolumn{11}{c}{9-shot} \\ 
\hline
Standard & 21.9 \small{$\pm$ 0.5}                      & 50.8 \small{$\pm$ 0.7}                      & 57.0 \small{$\pm$ 1.9}                      & 31.7 \small{$\pm$ 1.5}                      & 18.5 \small{$\pm$ 1.0}                      & 41.5 \small{$\pm$ 1.5}                      & 59.3 \small{$\pm$ 0.8}                      & 70.5 \small{$\pm$ 1.5}                     & 54.5 \small{$\pm$ 1.2}                      & 31.9 \small{$\pm$ 0.9} \\
Sorted+  & 22.3 \small{$\pm$ 0.9}                      & 51.4 \small{$\pm$ 1.1}                      & 55.3 \small{$\pm$ 1.3}                      & 32.1 \small{$\pm$ 0.8}                      & 18.8 \small{$\pm$ 3.5}                      & 41.2 \small{$\pm$ 0.6}                      & 58.9 \small{$\pm$ 1.2}                      & 71.1 \small{$\pm$ 1.7}                     & 55.6 \small{$\pm$ 0.5}                      & 30.5 \small{$\pm$ 0.6} \\
Sorted-  & 22.1 \small{$\pm$ 0.7}                      & 49.9 \small{$\pm$ 1.6}                      & 55.8 \small{$\pm$ 0.5}                      & 32.4 \small{$\pm$ 2.1}                      & 20.1 \small{$\pm$ 2.3}                      & 40.9 \small{$\pm$ 0.5}                      & 60.1 \small{$\pm$ 1.1}                      & 70.8 \small{$\pm$ 0.6}                     & 55.2 \small{$\pm$ 0.9}                      & 31.2 \small{$\pm$ 1.7} \\
{CBS}  & 18.5 \small{$\pm$ 1.3}                      & {52.1 \small{$\pm$ 1.8}}                      & 56.2 \small{$\pm$ 2.1}                      & 32.3 \small{$\pm$ 0.9}                      & 18.7 \small{$\pm$ 1.0}                      & 38.7 \small{$\pm$ 2.5}                      & 59.9 \small{$\pm$ 1.2}                      & 70.2 \small{$\pm$ 2.3}                     & 51.9 \small{$\pm$ 2.9}                      & 31.7 \small{$\pm$ 1.6} \\
{LENS}  & 21.5 \small{$\pm$ 1.0}                      & {52.3 \small{$\pm$ 1.4}}                      & 57.4 \small{$\pm$ 1.3}                      & 30.9 \small{$\pm$ 2.1}                      & 17.4 \small{$\pm$ 3.2}                      & 39.6 \small{$\pm$ 1.1}                      & 59.4 \small{$\pm$ 2.3}                      & 69.1 \small{$\pm$ 2.5}                     & 54.7 \small{$\pm$ 2.2}                      & 33.6 \small{$\pm$ 1.2} \\
PCW  & 23.5 \small{$\pm$ 1.2}                      & 51.7 \small{$\pm$ 0.8}                      & 54.9 \small{$\pm$ 0.7}                      & 32.8 \small{$\pm$ 1.6}                      & 19.6 \small{$\pm$ 1.7}                      & 41.3 \small{$\pm$ 0.9}                      & 61.5 \small{$\pm$ 1.6}                      & 71.3 \small{$\pm$ 1.2}                     & 52.1 \small{$\pm$ 1.4}                      & 32.8 \small{$\pm$ 1.1} \\
{SP}  & 23.4 \small{$\pm$ 1.4}                      & {53.1 \small{$\pm$ 1.5}}                      & 55.6 \small{$\pm$ 0.9}                      & 30.8 \small{$\pm$ 1.4}                      & 17.5 \small{$\pm$ 2.9}                      & 40.1 \small{$\pm$ 2.2}                      & 62.7 \small{$\pm$ 1.9}                      & 69.9 \small{$\pm$ 2.4}                     & 53.3 \small{$\pm$ 3.3}                      & 32.4 \small{$\pm$ 2.1} \\
ParaICL  & \textbf{\underline{25.4 \small{$\pm$ 0.8}}} & \textbf{\underline{54.3 \small{$\pm$ 0.9}}} & \textbf{\underline{58.1 \small{$\pm$ 2.3}}} & \textbf{\underline{33.9 \small{$\pm$ 1.7}}} & \underline{20.5 \small{$\pm$ 0.4}}          & \underline{42.8 \small{$\pm$ 0.5}}          & \textbf{\underline{63.2 \small{$\pm$ 0.9}}} & \textbf{\underline{72.9 \small{$\pm$ 1.3}}} & \underline{55.8 \small{$\pm$ 3.8}}         & \textbf{\underline{34.8 \small{$\pm$ 1.3}}} \\ 
\hline
\multicolumn{11}{c}{15-shot} \\ 
\hline
Standard & 21.2 \small{$\pm$ 1.0}                      & 50.3 \small{$\pm$ 0.9}                      & 54.8 \small{$\pm$ 3.6}                      & 29.0 \small{$\pm$ 2.2}                      & 18.1 \small{$\pm$ 1.3}                      & 40.4 \small{$\pm$ 0.5}                      & 61.9 \small{$\pm$ 2.7}                      & 60.6 \small{$\pm$ 4.7}                     & 49.9 \small{$\pm$ 2.5}             & 32.9 \small{$\pm$ 0.6} \\
Sorted+  & 22.4 \small{$\pm$ 0.8}                      & 49.8 \small{$\pm$ 0.6}                      & 55.4 \small{$\pm$ 2.1}                      & 30.1 \small{$\pm$ 1.5}                      & 19.3 \small{$\pm$ 0.9}                      & 41.7 \small{$\pm$ 1.6}                      & 62.5 \small{$\pm$ 1.3}                      & 60.3 \small{$\pm$ 2.1}                     & 52.7 \small{$\pm$ 0.9}             & 33.0 \small{$\pm$ 0.9} \\
Sorted-  & 21.8 \small{$\pm$ 1.1}                      & 50.9 \small{$\pm$ 1.7}                      & 55.2 \small{$\pm$ 1.8}                      & 30.6 \small{$\pm$ 2.9}                      & 18.7 \small{$\pm$ 0.6}                      & 39.4 \small{$\pm$ 0.7}                      & 62.1 \small{$\pm$ 2.1}                      & 63.5 \small{$\pm$ 3.4}                     & 51.2 \small{$\pm$ 1.8}             & 33.2 \small{$\pm$ 2.1} \\
{CBS}  & 20.1 \small{$\pm$ 1.0}                      & {50.2 \small{$\pm$ 0.8}}                      & 52.2 \small{$\pm$ 3.3}                      & 29.7 \small{$\pm$ 0.6}                      & 19.0 \small{$\pm$ 1.9}                      & 40.3 \small{$\pm$ 0.5}                      & 61.2 \small{$\pm$ 2.0}                      & 61.4 \small{$\pm$ 1.3}                     & 50.8 \small{$\pm$ 1.3}                      & 30.4 \small{$\pm$ 2.2} \\
{LENS}  & 21.3 \small{$\pm$ 2.3}                      & {48.5 \small{$\pm$ 3.5}}                      & 54.8 \small{$\pm$ 2.2}                      & 32.1 \small{$\pm$ 3.2}                      & 18.8 \small{$\pm$ 1.6}                      & 42.4 \small{$\pm$ 2.6}                      & 62.2 \small{$\pm$ 2.4}                      & 66.0 \small{$\pm$ 2.2}                     & 52.6 \small{$\pm$ 1.4}                      & 31.7 \small{$\pm$ 2.3} \\
PCW  & 23.8 \small{$\pm$ 0.9}                      & 49.7 \small{$\pm$ 1.4}                      & 55.1 \small{$\pm$ 0.9}                      & \underline{32.9 \small{$\pm$ 0.7}}                      & 19.9 \small{$\pm$ 1.1}                      & 41.8 \small{$\pm$ 0.3}                      & 62.7 \small{$\pm$ 1.5}                      & \underline{66.2 \small{$\pm$ 1.3}}                     & 50.1 \small{$\pm$ 0.7}                      & 32.8 \small{$\pm$ 0.8} \\
{SP}  & 22.6 \small{$\pm$ 1.4}                      & {50.1 \small{$\pm$ 1.1}}                      & 54.9 \small{$\pm$ 3.2}                      & 31.8 \small{$\pm$ 2.3}                      & 18.6 \small{$\pm$ 1.4}                      & 41.5 \small{$\pm$ 0.9}                      & 60.9 \small{$\pm$ 3.1}                      & 65.7 \small{$\pm$ 2.9}                     & 53.1 \small{$\pm$ 2.6}                      & 32.9 \small{$\pm$ 1.4} \\
ParaICL  & \underline{24.9 \small{$\pm$ 0.6}}          & \underline{51.0 \small{$\pm$ 2.1}}          & \underline{56.3 \small{$\pm$ 1.3}}          & {31.3 \small{$\pm$ 0.9}}          & \underline{20.6 \small{$\pm$ 0.5}}          & \textbf{\underline{42.9 \small{$\pm$ 1.1}}} & \textbf{\underline{63.2 \small{$\pm$ 3.2}}} & {65.8 \small{$\pm$ 0.8}}         & \underline{53.7 \small{$\pm$ 1.0}} & \underline{33.5 \small{$\pm$ 1.5}} \\ 
\bottomrule
\end{tabular}
}
\end{center}
\caption{Experimental results of Llama-2-7B-Chat and Mistral-7B-Instruct-v0.2 on various reasoning, natural language inference, and coding benchmarks. \underline{Underlined} indicates the highest scores for each shot group and \textbf{bold} indicates overall highest scores.}
\label{tab:main_exp}
\end{table*}

\begin{table}[tbh]
    \centering
        \resizebox{0.65\linewidth}{!}{
        \begin{tabular}{lcc}
        \toprule
        \textbf{Method} & \textbf{GSM8K} & \textbf{HellaSwag} \\
        \midrule
        Retrieval (5-shot) & 22.1 & 28.3 \\
        ParaICL (15-shot) & 24.9 & 31.3 \\
        \bottomrule
        \end{tabular}
        }
    \caption{ParaICL vs. retrieval-based ICL methods on GSM8K and HellaSwag.}
    \label{tab:retrievalicl}
\end{table}

\subsection{Datasets}
ParaICL is a versatile approach applicable to a broad range of tasks that can be framed into the few-shot ICL setting. 
We evaluate ParaICL extensively on several task categories, including reasoning, natural language inference (NLI), and coding.
For reasoning tasks, we evaluate on three datasets: 
(1) \textbf{GSM8K} \citep{cobbe2021gsm8k}, a mathematical reasoning dataset consisting of a collection of high-quality math word problems.
(2) \textbf{WinoGrande} \citep{sakaguchi2019winogrande}, a commonsense reasoning dataset of improved complexity beyond the Winograd Schema Challenge benchmark.
(3) \textbf{ARC} \citep{allenaiarc}, a knowledge reasoning dataset consisting of grad-school level science questions in a multiple-choice format. We adopt the challenge set of \textbf{ARC}.
For the NLI tasks, we select \textbf{HellaSwag} \citep{zellers2019hellaswag}, a commonsense NLI task that examines a model by predicting the most logical continuation of a described event.
In the coding category, we assess using \textbf{MBPP} \citep{austin2021mbpp}, a benchmark consisting of Python programming problems designed to be solvable by entry-level programmers.

\subsection{Baselines}
To provide a more comprehensive overview of where our framework stands, we use the following baselines:
(1) \textbf{Standard few-shot (Standard)} \citep{brown2020language}: Directly generating the results based on few-shot demonstration examples.
(2) \textbf{Semantically sorted few-shot (Sorted)} \citep{chen-etal-2023-many}: Utilizing the same few-shot demonstration examples, this approach organizes the examples by the semantic similarity between each example's question and the test question.
\textbf{Sorted+} indicates examples sorted in ascending similarity and \textbf{Sorted-} in descending similarity.
(3) \textbf{Coverage-based selection (CBS)} \citep{gupta-etal-2023-coverage}: CBS selects examples from more salient aspects such as reasoning patterns of the test input by using BERTScore-Recall.
(4) \textbf{Filter-then-search (LENS)} \citep{lens}: LENS utilizes InfoScore to evaluate and progressively selects examples.
(5) \textbf{Parallel context window (PCW)} \citep{ratner2023parallel}: Carving a long context into batches, PCW restricts the attention mechanism to apply within each batch. To ensure a fair comparison, we maintain the same number of batches as used in ParaICL.
(6) \textbf{Structured prompting (SP)} \citep{sp}: In SP, demonstration examples are separately encoded with well-designed position embeddings and jointly attended by the test example using a rescaled attention mechanism.

\subsection{Experiment Setup for ParaICL}
When setting up experiments for ParaICL, we maintain consistency with the demonstration examples used in baseline methods to ensure a fair comparison.
To demonstrate the effectiveness of ParaICL, we conduct experiments with varying numbers of total demonstration examples, specifically 3, 9, and 15, while keeping the count of parallel batches at three 
\footnote{ParaICL focuses on optimally using any number of available examples, and as such, we do not experiment with hundreds or thousands of examples.}.
Consequently, this results in 1, 3, and 5 demonstration examples per batch across the three settings.
The impact of batch numbers will be elaborated in Section \ref{sec:num_batches}.
Each experimental run is carried out with three random seeds for the selection of demonstration examples, and the results are averaged for reporting.
Our main experiments are conducted using open-source models\footnote{Due to cost concern, the validation of ParaICL's effectiveness on closed-source models is conducted on a more compact dataset in Section \ref{sec:close_llm}.}, including Llama-2-7B-Chat and Mistral-7B-Instruct-v0.2.
We utilize supervised SimCSE \citep{simcse} with BERT base to compute sentence embeddings.

\subsection{Experiment Results}
\label{sec:exp_results}
\paragraph{ParaICL consistently outperforms baseline methods}
We present the experimental results of all datasets from Llama-2-7B-Chat and Mistral-7B-Instruct-v0.2 in Table \ref{tab:main_exp}.
ParaICL consistently outperforms baseline methods on all datasets, demonstrating the effectiveness of our method.
On reasoning tasks, the average improvements with Llama-2-7B-Chat are 1.2\%, 2.7\%, and 2.0\% for 3-shot, 9-shot, and 15-shot settings, respectively.
Mistral-7B-Instruct-v0.2 shows improvements of 1.0\%, 2.9\%, and 3.0\% for the same settings, demonstrating ParaICL's adaptability across different large language models.
On commonsense NLI and coding tasks, ParaICL presents an average performance boost of 1.8\% and {2.0\%} for Llama-2-7B-Chat, and 1.9\% and {2.0\%} for Mistral-7B-Instruct-v0.2, respectively.
Moreover, ParaICL consistently surpasses CBS and PCW in all settings and on virtually all datasets, underscoring the advantages of utilizing semantic information in our approach.
An observation is made regarding the performance similarity between the Sorted+ and Sorted- methods.
Sorted+ sometimes outperforms Sorted- and vice versa.
This finding is consistent with prior research by \citet{levy2023diverse} and \citet{liu2021makes}.
With ParaICL, this issue is lessened due to the reduced number of demonstration examples per batch.

\paragraph{Number of demonstration examples}
As shown in Table \ref{tab:main_exp}, ParaICL consistently exhibits enhanced performance across multiple datasets under 3-shot, 9-shot, and 15-shot settings.
However, the performance gain in the 3-shot setting is less marked compared to the 9-shot and 15-shot settings. 
This variation is attributed to the fact that the quantity of demonstration examples in each batch sets a limit on ParaICL's maximum performance. 
It is also observed that on certain datasets, an increase in demonstration examples can actually degrade performance. 
For instance, in the HellaSwag dataset, the Standard method's performance decreases as the number of shots increases: 32.9\% for 3-shot, 31.7\% for 9-shot, and 29.0\% for 15-shot. 
This decline is likely due to potential label bias within the demonstration examples, a phenomenon noted in various studies \citep{wang2023selfconsistency}. 
Additionally, HellaSwag demands commonsense knowledge.
Earlier research has indicated that merely increasing the number of demonstrations does not guarantee improved the performance \citep{yao2023react, li2024cok}.

\paragraph{ParaICL vs. retrieval-based ICL methods}
Retrieval-based ICL methods are designed to select the most relevant demonstration examples from a pool of candidates, in contrast, ParaICL utilizes all available demonstration examples.
We compare the performance of ParaICL with retrieval-based ICL methods in a 15-shot setting on the GSM8K and HellaSwag datasets.
Retrieval-based ICL methods are restricted to choosing the optimal demonstration examples based on sentence similarity \citep{liu2021makes} from a set of 15 candidates.
As shown in Table \ref{tab:retrievalicl}, ParaICL surpasses the retrieval-based method on both GSM8K and HellaSwag datasets.
The necessity for a large candidate pool to select from, which retrieval-based ICL methods rely on, hampers their adaptability when only a limited number of demonstration examples are available.
ParaICL, however, is capable of efficiently utilizing any quantity of demonstration examples, showcasing its superior versatility.

\section{Ablation Studies and Analysis}

\begin{figure}[t!]
    \centering
    \includegraphics[width=0.9\linewidth]{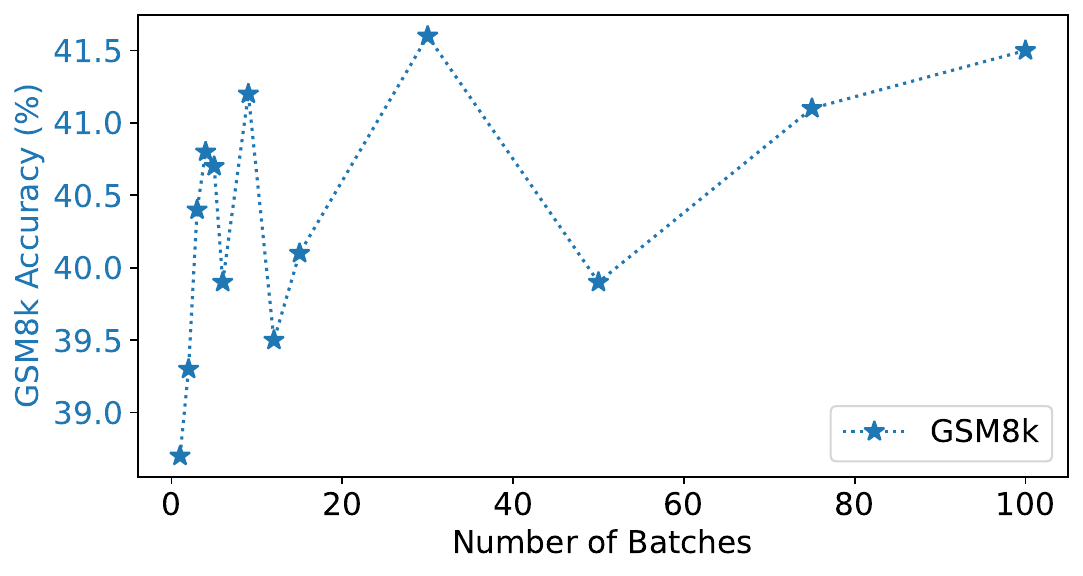}
    \caption{Results of Mistral-7B-Instruct-v0.2 on GSM8K using different batches of five-shot demonstration examples.}
    \label{fig:ab_num_batch}
\end{figure}

\subsection{Number of Batches}
\label{sec:num_batches}
ParaICL keeps the context length for each batch manageable, allowing for the increasing of batch numbers as hardware capabilities allow.
In this experiment, we set the number of demonstration examples in each batch as five, and progressively increase the number of batches.
The results of Mistral-7B-Instruct-v0.2 on GSM8K is in Figure~\ref{fig:ab_num_batch}.
With the increment in batch numbers, ParaICL's improvements tend to converge at five batches.
Further increase in batch numbers leads to instability in results.
We attribute this to the higher likelihood of incorporating batches that negatively affect token selection.
Consequently, we set the batch number to three in our main experiments to ensure stable performance improvements.

\begin{table}[t!]
    \centering
        \resizebox{0.65\linewidth}{!}{
        \begin{tabular}{lcc}
        \toprule
        \textbf{Method} & \textbf{GSM8K} & \textbf{HellaSwag} \\
        \midrule
        w/o S.S. (9-shot) & 21.1 & 32.2 \\
        w/ S.S. (9-shot) & 25.4 & 33.9 \\
        \bottomrule
        \end{tabular}
        }
    \caption{With vs. without semantic sorting on GSM8K and HellaSwag. S.S. stands for semantic sorting.}
        \label{tab:semanticsorting}
\end{table}

\subsection{Parallel Batching with vs. without Semantic Sorting}
The demonstration examples are sequenced by their question semantic similarities to the test question before parallel batching.
This ensures that semantically similar examples are batched together, allowing them to share the same batch similarity score during decoding.
According to results shown in Table~\ref{tab:semanticsorting}, parallel batching with semantic sorting significantly outperforms without semantic sorting.
In fact, Parallel batching without semantic sorting even performs worse than the standard approach on GSM8K.
This is because, without semantic sorting, a single batch can contain both relevant and irrelevant demonstration examples.
This mix of examples reduces the overall decoding probability from the relevant examples in the batch.

\begin{table}[t]
    \centering
        \resizebox{0.6\linewidth}{!}{
        \begin{tabular}{lcc}
        \toprule
        \textbf{Method} & \textbf{Llama-2} & \textbf{Mistral} \\
        \midrule
        w/ $\mathcal{V}_{head}$ (9-shot) & 25.4 & 42.8 \\
        w/o $\mathcal{V}_{head}$ (9-shot) & 2.9 & 3.1 \\
        \bottomrule
        \end{tabular}
        }
    \caption{Results of ParaICL on GSM8K with and without $\mathcal{V}_{head}$.}
    \label{tab:apc}
\end{table}

\begin{table*}[t!]
    \centering
        \resizebox{0.86\linewidth}{!}{
        \begin{tabular}{lcccc}
        \toprule
        \textbf{Method} & \textbf{Llama-2-13B-Chat} & \textbf{Llama-3.1-70B-Instruct} & \textbf{Mixtral-8x7B-Instruct-v0.1} &\textbf{\texttt{gpt-3.5-turbo-instruct}} \\
        \midrule
        Standard (9-shot) & 23.1 & 86.4 & 58.1 & 62.0 \\
        ParaICL (9-shot) & 25.6 & 87.1 & 59.2 & 66.0 \\
        \bottomrule
        \end{tabular}
        }
    \caption{Experimental results of ParaICL on GSM8K using more open- and closed-source LLMs.}
        \label{tab:chatgpt}
\end{table*}

\begin{table}[t!]
    \centering
        \resizebox{0.75\linewidth}{!}{
        \begin{tabular}{lcc}
        \toprule
        \textbf{Method} & \textbf{Llama-2} & \textbf{Mistral} \\
        \midrule
        Standard (9-shot) & 21.9 & 41.5 \\
        ParaICL (9-shot) & 25.4 & 42.8 \\
        ParaICL w. C.D. (9-shot) & 26.1 & 43.1 \\\hline
        Standard (15-shot) & 21.1 & 40.4 \\
        ParaICL (15-shot) & 24.9 & 42.9 \\
        ParaICL w. C.D. (15-shot) & 25.2 & 43.5 \\
        \bottomrule
        \end{tabular}
        }
    \caption{Results of ParaICL with contrastive decoding on GSM8K. C.D. stands for contrastive decoding.
        }
        \label{tab:integrate}
\end{table}

\subsection{Adaptive Plausibility Constraint}
The adaptive plausibility constraint plays a crucial role in our approach, akin to its importance in contrastive decoding as noted by \citet{li2023contrastive}. 
To assess the impact of this constraint, we carried out an ablation study by eliminating it from our method. The outcomes clearly show a substantial decline in performance, as detailed in Table~\ref{tab:apc}. 
This observation is consistent with findings from the contrastive decoding paper.

\subsection{{More Open- and Closed-Source LLMs}}
\label{sec:close_llm}

In this section, we demonstrate that ParaICL is effective in both open- and closed-source LLMs.

We first conduct experiments on GSM8K using three additional open-source models spanning a range of parameter counts, including Llama-2-13B-Chat, Llama-3.1-70B-Instruct, and Mixtral-8x7B-Instruct-v0.1.
As shown in Table~\ref{tab:chatgpt}, ParaICL consistently enhances the performances, highlighting its adaptability to open-source models.

Our method necessitates the next token probabilities from the model (\textit{i.e.,} the softmax of the raw scores generated by the final layer) for computing the semantic weighted average.
However, closed-source LLMs, due to their proprietary nature, do not make these probabilities available.
For instance, OpenAI's \texttt{gpt-3.5-turbo-instruct} model only allows access to a maximum of five tokens that have top log probabilities, which represents the extent of information available for our use.
We employ the same steps as shown in Section~\ref{sec:psd} to execute parallel semantic decoding using the provided log probabilities from OpenAI models.
Concerns regarding API costs lead us to limit our experimentation to a randomly chosen set of 50 data points from the GSM8K datasets.
Table~\ref{tab:chatgpt} illustrates that ParaICL enhances performance beyond the standard method, further showcasing its adaptability even with limited information on token distributions.
In contrast, PCW is restricted to open-source models as it requires modifications to the attention mask.

\subsection{Integration with Other Methods}
\label{sec:integration}
In this section, we demonstrate that ParaICL is compatible with other methods.
Specifically, we explore its integration with contrastive decoding (CD) \citep{li2023contrastive}.
Building upon the concept of contrastive objectives introduced by \citet{li2023contrastive}, which leverages the differential signals between larger and smaller language models for decoding, we incorporate contrastive batches into ParaICL.
This integration involves calculating the weighted average distributions for both positive and contrastive batches individually, then applying the contrastive objective by subtracting the logarithmic values of these two distributions.
This process helps in selecting tokens generated from the positive batches that are least similar to those from the contrastive batches, thus refining the selection for more plausible outcomes.
We experiment on GSM8K.
We adopt the contrastive chain-of-thought as outlined by \citet{chia2023contrastive}, creating a batch consisting of five reasoning failure cases.
These cases encompass invalid reasoning, incoherent objects, incoherent language, irrelevant objects, and irrelevant language.
The specifics of these demonstration examples are provided in Appendix~\ref{app:contrastive}.
The integration of ParaICL with CD has shown to enhance performance on both the Llama-2-7B-Chat and Mistral-7B-Instruct-v0.2 models, particularly in 9- and 15-shot settings, as illustrated in Table~\ref{tab:integrate}. 
This further evidences the flexibility and improved effectiveness of ParaICL when combined with other methodologies.


\subsection{Majority Voting vs. Standard Average vs. Weighted Average}
\label{sec:majority_voting}

\begin{table}[t!]
    \centering
        \resizebox{0.75\linewidth}{!}{
        \begin{tabular}{lcc}
        \toprule
        \textbf{Method} & \textbf{GSM8K} & \textbf{HellaSwag} \\
        \midrule
        Majority Voting (9-shot) & 20.3 & 32.7 \\
        Standard average (9-shot) & 23.6 & 32.8 \\
        Weighted average (9-shot) & 25.4 & 33.9 \\
        \bottomrule
        \end{tabular}
        }
    \caption{Majority Voting vs. Weighted Average on GSM8K and HellaSwag.}
        \label{tab:majority}
\end{table}

As aforementioned in Section~\ref{sec:psd}, demonstration examples have varying semantic contribution to the test sample.
As such, we utilize weighted average semantic objective during the parallel semantic decoding.
We study the effectiveness of the weighted average method by comparing it with the majority voting and standard average methods.
According to the results presented in Table~\ref{tab:majority}, utilizing weighted average during parallel semantic decoding substantially surpasses the performance achieved through majority voting and standard average.
Notably, the application of majority voting on the GSM8K dataset results in performance that is even worse than the standard method.
This can be attributed to irrelevant batches contributing votes that may dominate over the more desired tokens.
In contrast, the weighted average method ensures that batches with the highest relevance have the greatest impact on the selection of subsequent tokens, leading to more accurate outcomes.

\subsection{Performance-Efficiency Tradeoff}
ParaICL introduces additional inference time due to the computation of embeddings for both the demonstrations and test examples.
Nevertheless, this overhead is minimal when computations are conducted in parallel.
Across various datasets, the average time to compute embeddings using SimCSE \citep{simcse} is merely 5.1 seconds.

\section{Conclusions}
In this work, we introduce a novel approach known as parallel in-context learning (ParaICL), designed to enhance the effectiveness of few-shot in-context learning.
ParaICL aims to fully leverage all available demonstration examples while keeping within the limits of a manageable input context size. 
It starts by executing parallel batching, grouping demonstration examples into various batches based on the semantic similarities between the questions in the demonstrations and the test questions. 
Afterward, for each batch, normalized semantic scores are calculated. 
The process culminates in the decoding of the final tokens, optimizing a weighted average semantic objective under an adaptive plausibility constraint. 
ParaICL has been proven to yield consistent performance improvements across a diverse set of benchmarks and exhibits a high degree of compatibility for integration with other methodologies.

\section*{Limitations}
We only study the effectiveness of ParaICL in decoder-only language models.
In future works, we aim to extend our evaluation of ParaICL to language models that utilize different structures, including those with encoder-only and encoder-decoder structures. 
In this paper, we only demonstrate that ParaICL can be easily integrated with contrastive objectives.
We would like to explore the integration of ParaICL with other methods, such as retrieval-augmented generation.

\section*{Ethical Impact}
We do not foresee any potential ethical issues with our proposed method.

\section*{Acknowledgements}
This work was supported by DAMO Academy through DAMO Academy Research Intern Program.

\bibliography{custom}

\appendix

\section{Appendix}
\label{sec:appendix}
\subsection{Datasets}
GSM8K \citep{cobbe2021gsm8k} and HellaSwag \citep{zellers2019hellaswag} are under the MIT License.
WinoGrande \citep{sakaguchi2019winogrande} and MBPP \citep{austin2021mbpp} are under the Apache-2.0 License.
ARC \citep{allenaiarc} is under the GPL-3.0 License.

\subsection{Contrastive demonstration examples for GSM8K}
\label{app:contrastive}
\paragraph{Invalid reasoning:}
\textbf{Question:} Janet's ducks lay 16 eggs per day. She eats three for breakfast every morning and bakes muffins for her friends every day with four. She sells the remainder at the farmers' market daily for \$2 per fresh duck egg. How much in dollars does she make every day at the farmers' market?
\textbf{Answer:} Janet sells 16 - 3 - 4 = \textless{}\textless{}16-3-4=9\textgreater{}\textgreater{}9 duck eggs a day. She makes 9 + 2 = \$\textless{}\textless{}9+2=11\textgreater{}\textgreater{}11 every day at the farmer's market. \#\#\#\# 11

\paragraph{Incoherent objects:}
\textbf{Question:} Janet's ducks lay 16 eggs per day. She eats three for breakfast every morning and bakes muffins for her friends every day with four. She sells the remainder at the farmers' market daily for \$2 per fresh duck egg. How much in dollars does she make every day at the farmers' market?
\textbf{Answer:} Janet sells 9 * 2 = \textless{}\textless{}9*2=18\textgreater{}\textgreater{}18 duck eggs a day. She makes 16 - 3 - 4 = \$\textless{}\textless{}16-3-4=9\textgreater{}\textgreater{}9 every day at the farmer's market.\#\#\#\# 18

\paragraph{Incoherent language:}
\textbf{Question:} Janet's ducks lay 16 eggs per day. She eats three for breakfast every morning and bakes muffins for her friends every day with four. She sells the remainder at the farmers' market daily for \$2 per fresh duck egg. How much in dollars does she make every day at the farmers' market?
\textbf{Answer:} Janet sells 16 - 3 - 4 = \textless{}\textless{}16-3-4=9\textgreater{}\textgreater{}9 duck eggs a day.She makes 9 - 2 = \$\textless{}\textless{}9-2=7\textgreater{}\textgreater{}7 every day at the farmer's market.\#\#\#\# 7

\paragraph{Irrelevant objects:}
\textbf{Question:} Janet's ducks lay 16 eggs per day. She eats three for breakfast every morning and bakes muffins for her friends every day with four. She sells the remainder at the farmers' market daily for \$2 per fresh duck egg. How much in dollars does she make every day at the farmers' market?
\textbf{Answer:} Janet sells 17 - 5 - 6 = \textless{}\textless{}17-5-6=6\textgreater{}\textgreater{}6 duck eggs a day. She sells 6 * 3 = \$\textless{}\textless{}6*3=18\textgreater{}\textgreater{}18 every day at the duck eggs.\#\#\#\# 18

\paragraph{Irrelevant language:}
\textbf{Question:} Janet's ducks lay 16 eggs per day. She eats three for breakfast every morning and bakes muffins for her friends every day with four. She sells the remainder at the farmers' market daily for \$2 per fresh duck egg. How much in dollars does she make every day at the farmers' market?
\textbf{Answer:} Janet sells 16 - 3 - 4 = \textless{}\textless{}16-3-4=9\textgreater{}\textgreater{}9 duck eggs a day. She wants her hair to be 9 * 2 = \$\textless{}\textless{}9*2=18\textgreater{}\textgreater{}18 inches long when she cuts it.\#\#\#\# 18



\end{document}